\documentclass[runningheads]{llncs}
\usepackage[T1]{fontenc}
\usepackage[dvipdfmx]{graphicx}
\usepackage{booktabs}
\usepackage[table]{xcolor}
\usepackage{multirow}
\usepackage{enumitem}
\usepackage{hyperref}
\begin{document}
\title{Knowledge Distillation Framework for Accelerating High-Accuracy Neural Network-Based Molecular Dynamics Simulations}
\author{Naoki Matsumura \and 
Yuta Yoshimoto \and
Yuto Iwasaki \and
Meguru Yamazaki \and
Yasufumi Sakai}
\authorrunning{N. Matsumura et al.}
\institute{Fujitsu Research, Fujitsu Limited, 4-1-1, Kamiodanaka, Nakahara-ku, Kawasaki, Kanagawa 211-8588, Japan\\
\email{matsumura-naoki@fujitsu.com}}
\maketitle              
\begin{abstract}
Neural network potentials (NNPs) offer a powerful alternative to traditional force fields for molecular dynamics (MD) simulations. Accurate and stable MD simulations, crucial for evaluating material properties, require training data encompassing both low-energy stable structures and high-energy structures. Conventional knowledge distillation (KD) methods fine-tune a pre-trained NNP as a teacher model to generate training data for a student model. However, in material-specific models, this fine-tuning process increases energy barriers, making it difficult to create training data containing high-energy structures. To address this, we propose a novel KD framework that leverages a non-fine-tuned, off-the-shelf pre-trained NNP as a teacher. Its gentler energy landscape facilitates the exploration of a wider range of structures, including the high-energy structures crucial for stable MD simulations. Our framework employs a two-stage training process: first, the student NNP is trained with a dataset generated by the off-the-shelf teacher; then, it is fine-tuned with a smaller, high-accuracy density functional theory (DFT) dataset. We demonstrate the effectiveness of our framework by applying it to both organic (polyethylene glycol) and inorganic (L$_{10}$GeP$_{2}$S$_{12}$) materials, achieving comparable or superior accuracy in reproducing physical properties compared to existing methods. Importantly, our method reduces the number of expensive DFT calculations by 10x compared to existing NNP generation methods, without sacrificing accuracy. Furthermore, the resulting student NNP achieves up to 106x speedup in inference compared to the teacher NNP, enabling significantly faster and more efficient MD simulations.

\keywords{Knowledge Distillation \and Neural Network Potential \and Molecular Dynamics}
\end{abstract}
%
\section{Introduction}
Neural network potentials (NNPs) represent a significant advancement in interatomic potential modeling, offering a data-driven alternative to traditional empirical force fields. By leveraging the representational power of neural networks, NNPs learn the complex potential energy surface (PES) directly from high-accuracy ab initio calculations, such as density functional theory (DFT). This approach overcomes the limitations of fixed functional forms inherent in empirical potentials, enabling accurate molecular dynamics (MD) simulations and rapid materials screening~\cite{unke_machine_2021,friederich_machine-learned_2021}.

To further enhance the prediction accuracy of NNPs, researchers have increasingly focused on developing larger and more complex models. 
As a result, the learning of large datasets has become possible, leading to the development of universal NNPs applicable to a wide range of materials~\cite{chen_universal_2022,deng_chgnet_2023,yang_mattersim_2024}. 
While these efforts have led to improved accuracy and versatility, they have also resulted in a significant increase in computational cost. This slowdown in inference speed limits the applicability of these large NNPs in computationally demanding applications, such as large-scale, long-time MD simulations.
This highlights the growing need for developing NNPs that are both highly accurate and computationally efficient.

\begin{figure}[t]
    \includegraphics[width=\textwidth]{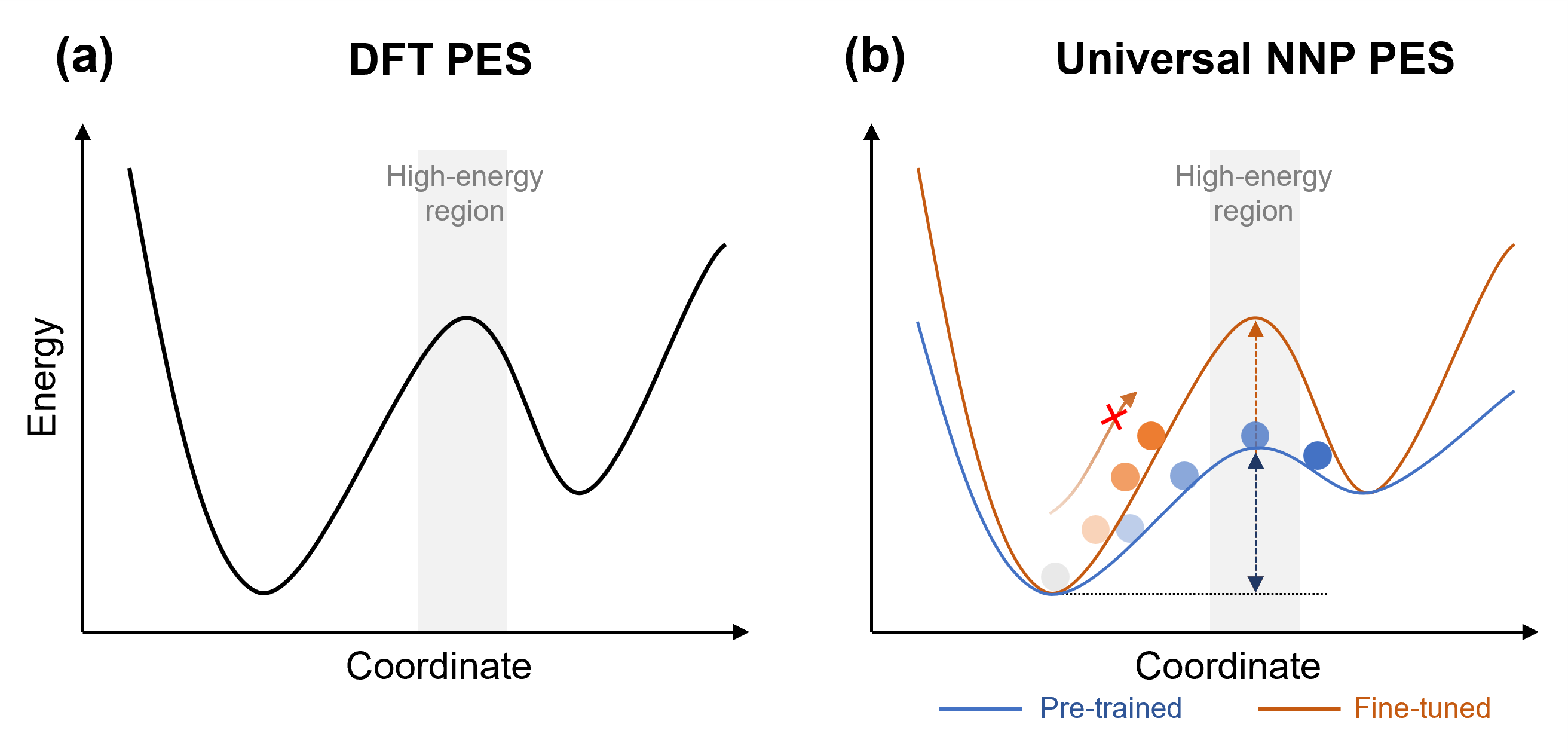}
    \caption{Illustration of the potential energy surface (PES). (a) PES calculated from DFT calculations and (b) PES of pre-trained universal NNP (blue) and a DFT fine-tuned universal NNP (orange). While fine-tuning with DFT data enhances the reproducibility of the DFT PES, especially in the high-energy region, it can lead to a reduced exploration of structures in the high-energy region.}
    \label{fig:PES}
\end{figure}

To address this limitation, knowledge distillation (KD)~\cite{hinton_distilling_2015} has emerged as a promising technique. KD transfers knowledge from a larger, accurate ``teacher'' model to a smaller ``student'' model, enabling faster inference speeds. In KD, soft targets refer to the teacher model's outputs that reflects information of the teacher model, while hard targets refer to the ground truth labels.
Several KD methods have been proposed for NNPs in MD simulations~\cite{zhang_dpa-2_2024,wang_pfd_2025}. These methods typically begin by fine-tuning a pre-trained universal NNP with DFT data to create a teacher model that accurately reproduces the DFT PES (Figure~\ref{fig:PES}(a)). The teacher model is then used to generate a dataset of structures for training a student model. 
As illustrated in Figure~\ref{fig:PES}(b), pre-trained universal NNPs are known to underestimate the energies in high-energy regions~\cite{deng_systematic_2025}. 
Indeed, fine-tuning the teacher model allows the shape of PES to approach that of DFT, enabling accurate prediction of structures in high-energy regions. However, this also increases the energy barriers in MD simulations, making it difficult to explore structures in high-energy regions (crucial for improving the robustness of the NNP, as demonstrated in previous works~\cite{matsumura_generator_2025,yoshimoto_large-scale_2025}).
This effect is visually represented in Figure~\ref{fig:PES}(b), where the orange curve, representing the PES of fine-tuned universal NNP, exhibits higher energy barriers compared to that of pre-trained universal NNP.
From the perspective of exploring a wide range of structures, including those in high-energy regions, using a non-fine-tuned, off-the-shelf teacher model with a gentler shape of PES is beneficial because it lowers the energy barriers required to reach high-energy regions.

Here, we present a novel KD framework for generating high-accuracy and computationally efficient NNPs to realize stable MD simulation. To effectively explore structures in high-energy regions, we leverage a non-fine-tuned, off-the-shelf pre-trained NNP as a teacher. Its gentler energy landscape facilitates the exploration of a wider range of structures, including the high-energy structures crucial for stable MD simulations.
Our framework employs a two-stage training process: first, the student model is trained with a dataset generated by the off-the-shelf teacher model; then, it is fine-tuned with a smaller, high-accuracy DFT dataset collected by the efficient structural feature-based screening~\cite{matsumura_generator_2025}, to further refine the student model's accuracy.

In this work, we demonstrate the effectiveness of our proposed framework by applying it to both organic and inorganic materials: polyethylene glycol (PEG) and L$_{10}$GeP$_2$S$_{12}$ (LGPS), respectively. We perform comprehensive validation to assess the accuracy and efficiency of the student NNPs. For PEG, we compare our framework with an existing KD method to highlight its advantages. Finally, we evaluate the computational performance of both the teacher and student NNPs, and we compare the NNP generation time with an existing active learning-based method to demonstrate the efficiency of our proposed framework.

Our contributions to the knowledge distillation of NNPs for MD simulations are summarized as follows:  
\begin{itemize}[label=•]
\item We propose a novel knowledge distillation framework for generating high-accuracy and computationally efficient NNPs to realize long-time MD simulation of large-scale material systems. 
\item Our key proposal is training a student model with soft targets generated by a non-fine-tuned, off-the-shelf teacher model, enabling the generation of diverse structures in high-energy regions and more robust MD simulation.
\item We demonstrate that our proposed framework achieves comparable or superior accuracy in reproducing experimental properties of PEG and LGPS compared to existing methods.
\item Our framework reduces the number of hard targets generated using expensive DFT calculations by 10x compared to existing NNP generation methods, without sacrificing accuracy.
\end{itemize}

\section{Related work}
\subsubsection{Molecular simulations with NNPs.}
In this work, we focus on atomic systems characterized by atomic numbers, positions and velocities. Our goal is to develop an NNP model capable of predicting the energy and forces acting on each atom, crucial properties for determining system stability and enabling molecular dynamics simulations. Typically, an NNP model consists of two key neural networks: a descriptor and fitting networks~\cite{behler_generalized_2007,zhang_end--end_2018}. The descriptor network transforms the atomic environment of each atom into a set of numerical features that are invariant to translation, rotation, and permutation of atoms. These features are then input to the fitting network, which maps the features to the potential energy and atomic forces. By accurately learning this mapping, NNPs can provide a computationally efficient alternative to \textit{ab initio} methods for simulating complex molecular systems~\cite{unke_machine_2021,friederich_machine-learned_2021}.

\subsubsection{Generating NNPs: From \textit{ab initio} MD to universal models.}
NNPs are typically trained on datasets of atomic structures with corresponding energies and forces. One common approach for obtaining labeled atomic structures is to employ \textit{ab initio} MD (AIMD)~\cite{chahal_deep-learning_2024,liu_compression_2024}. However, AIMD can be inefficient as it generates a large number of similar structures, leading to redundant DFT calculations that contribute little to the NNP's learning process. To overcome this, active learning (AL) methods have emerged as a promising alternative for efficient data acquisition~\cite{zhang_dp-gen_2020,matsumura_generator_2025,yoshimoto_large-scale_2025}, aiming to generate high-accuracy NNPs with strategically collected data. 
Matsumura et al. and Yoshimoto et al.~\cite{matsumura_generator_2025,yoshimoto_large-scale_2025} have shown that including the configurations within high-energy region is crucial for improving NNP robustness. This enhanced robustness enables more reliable and accurate simulations, enabling large-scale and long-time NNP-MD simulations.
Nevertheless, these approaches still require a large amount of data, and the cost associated with iterative training process can be noticeable. More recently, universal NNP models, trained on vast datasets encompassing a wide range of elements, such as M3GNet~\cite{chen_universal_2022}, CHGNet~\cite{deng_chgnet_2023} and MatterSim~\cite{yang_mattersim_2024}, have emerged. MatterSim was trained on 17 million structures from MPF2021~\cite{chen_universal_2022}, MPtrj~\cite{deng_chgnet_2023}, Alexandria~\cite{schmidt_machine-learning-assisted_2023}, and their own dataset~\cite{zeni_generative_2025}, enabling stable and high-accuracy NNP-MD simulation for a wide variety of systems. These models learn general interatomic interactions and can be transfer-learned for specific materials~\cite{yang_mattersim_2024,deng_systematic_2025}. However, their slow inference speeds and large memory footprints, stemming from their underlying graph neural network (GNN) architecture, limit their use in large-scale simulations or resource-constrained environments~\cite{wines_chips-ff_2024,jacobs_practical_2025}.

\subsubsection{Knowledge distillation.}
Knowledge distillation (KD)~\cite{hinton_distilling_2015} is a technique for model compression and acceleration, finding widespread application in domains such as computer vision~\cite{wang_knowledge_2022} and natural language processing~\cite{gupta_compression_2022,yang_survey_2024}. The main objective of KD is transferring knowledge from a large and complex teacher model to a lightweight and fast student model. 
While various KD approaches exist, including those leveraging intermediate feature representations as hints~\cite{romero_fitnets_2014} or matching the flow of solution procedure (FSP) matrices between teacher and student~\cite{yim_gift_2017}, this work adopts the straightforward method proposed by Hinton et al.~\cite{hinton_distilling_2015}, where the student directly learns from the teacher's output.
Within the realm of NNPs, Kelvinius et al.~\cite{kelvinius_accelerating_2023} introduced node-to-node, edge-to-edge, and vector-to-vector KD strategies to transfer knowledge from lager GNNs to smaller ones, demonstrating improved accuracy in the student model. 
One of the challenges faced in their work is the substantial computational cost associated with the need for large DFT datasets, which can often reach into the millions. In our work, we offer a KD approach that reduces the number of required DFT data points to a more manageable scale, on the order of thousands. We note that their method are designed for learning across multiple systems simultaneously, which differs from our focus on KD tailored for MD simulations.
Zhang et al.~\cite{zhang_dpa-2_2024} and Wang et al.~\cite{wang_pfd_2025} proposed a KD method for generating NNPs for MD simulations. In this method, they first fine-tune a pre-trained universal NNP with DFT data to construct a teacher model. This teacher model is subsequently employed to produce soft targets for training the student model.
It is well-documented that pre-trained universal NNPs tend to underestimate energies in high-energy regions~\cite{deng_systematic_2025}. Fine-tuning the teacher model effectively adjusts the shape of PES to align with DFT, thereby enhancing the accuracy of predictions for structures in high-energy regions. However, this adjustment also elevates the energy barriers in MD simulations, hindering the exploration of structures in high-energy regions, which are crucial for improving robustness of NNPs~\cite{matsumura_generator_2025,yoshimoto_large-scale_2025}.
Therefore, different from the method, we use a non-fine-tuned, off-the-shelf teacher model with a gentler shape of PES to generate soft targets, as it reduces the energy barriers needed to access high-energy regions.

\section{Knowledge Distillation Framework}
In this section, we introduce a KD framework to construct a high-accuracy and computationally efficient NNP models specialized for MD, from a pre-trained universal NNP. Hereafter, we refer to the universal NNP as the teacher model, and the distilled NNP as the student model.
An overview of the proposed framework is shown in Figure~\ref{fig:distillation_process}.
Unlike existing methods~\cite{zhang_dpa-2_2024,wang_pfd_2025}, our framework begins by generating soft targets using a pre-trained teacher model. This approach contribute to generate structures within high-energy regions, which is crucial for improving robustness of NNPs~\cite{matsumura_generator_2025,yoshimoto_large-scale_2025}. Subsequently, to enhance of the reproducibility of the student NNP PES to DFT PES, we generate hard targets and fine-tune the trained student model. These hard targets are labeled by DFT calculations performed on strategically selected structures derived from the soft targets.
Each process is described in detail below:

\begin{figure}[t]
    \includegraphics[width=\textwidth]{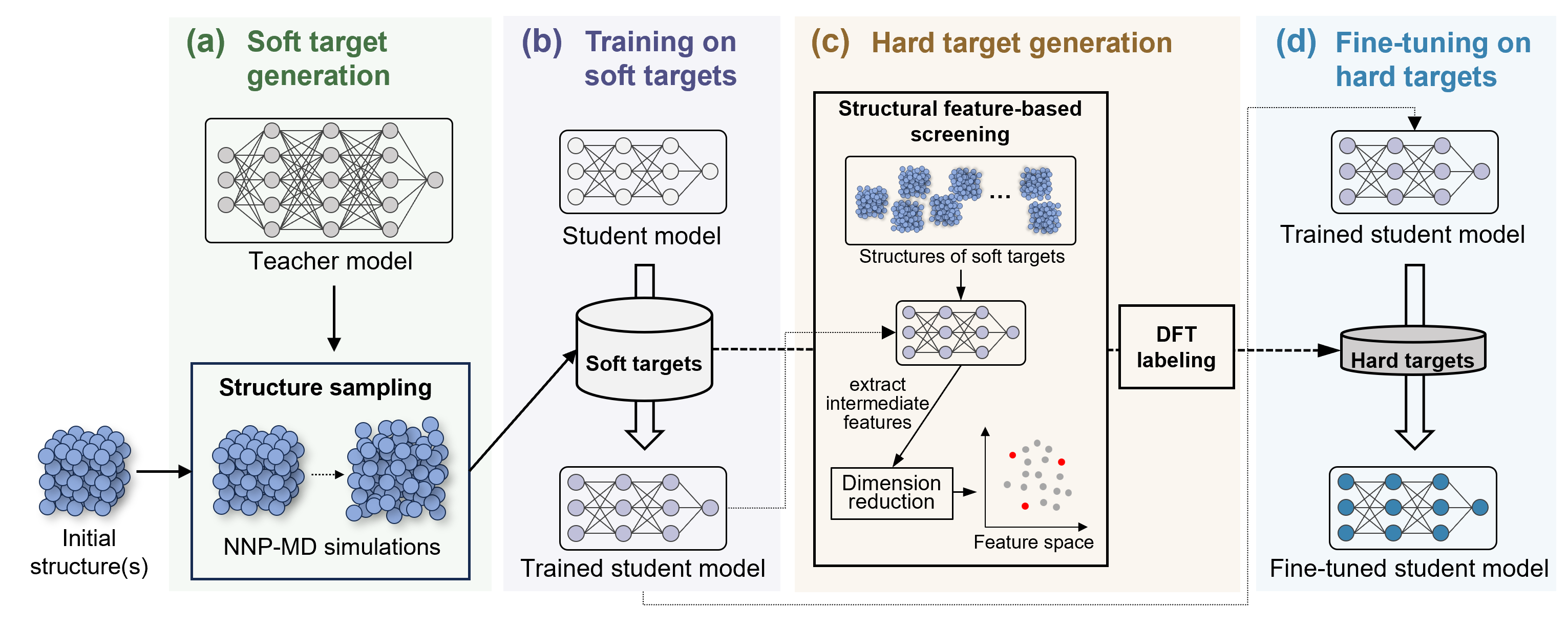}
    \caption{Overview of the proposed knowledge distillation framework. (a) NNP-MD simulations are performed on initial structures using a non-fine-tuned, off-the-shelf teacher model to generate soft targets. (b) A student model is then trained on the soft targets. (c) Data points are selected from the soft targets using a structural feature-based screening method~\cite{matsumura_generator_2025} and converted into hard targets using DFT calculations. (d) The student model trained on soft targets is fine-tuned on hard targets.}
    \label{fig:distillation_process}
\end{figure}

\subsubsection{(a) Soft target generation:}
First, we perform NNP-MD simulations using a teacher model to generate a comprehensive set of atomic structures. To effectively sample structures within high-energy regions, our proposed framework leverages the off-the-shelf teacher model for structural exploration. The NNP-MD simulation parameters (e.g., temperature, pressure, and simulation time) are carefully chosen to cover the conditions relevant to the downstream task. Subsequently, these generated structures are labeled with soft targets, including energy and force. The soft targets reflect the teacher model's knowledge of the PES and guide the initial learning phase of the student model.

\subsubsection{(b) Learning with soft targets:}
A student model is trained using the soft targets produced in step (a). This stage aims to train the student model on the PES of the teacher model and to enrich the descriptors of the student model. Following the observation by Kelvinius et al.~\cite{kelvinius_accelerating_2023}, the application of Kullback-Leibler divergence loss, which is often advantageous in classification tasks, provides limited gains in the context of NNP modeling due to the regression-based output~\cite{saputra_distilling_2019}. Consequently, we employ mean squared error (MSE) loss in this work. The resulting trained student model functions as a structural selector for generating hard targets in the subsequent step (c), and as a pre-trained model for fine-tuning in step (d).

\subsubsection{(c) Hard target generation:}
To further improve the accuracy of the student model, we generate hard targets, which offer higher precision than the soft targets. This involves selecting a representative subset of structures from the larger pool of soft targets and converting them into hard targets using DFT calculations. Considering the computational expense of DFT calculations, efficient data point selection is crucial. Therefore, we employ the structural feature-based screening method, used in the GeNNIP4MD software~\cite{matsumura_generator_2025}, for efficient data point selection. In this screening method, the atomic structure of each soft target is input into the student model, trained in step (b), and intermediate layer features are extracted. These features are then compressed into a two-dimensional (2D) feature space using a dimensionality reduction technique such as densMAP~\cite{narayan_assessing_2021}. In addition to feature diversity, energy diversity is also crucial for generating robust NNP models~\cite{qi_robust_2024,kaplan_foundational_2025}. To account for this, we augment the 2D feature space with a third dimension representing normalized energy, creating a three-dimensional (3D) space.
Finally, data points are selected in this 3D space by maximizing the inter-point distances, and DFT calculations are performed on these selected points to generate the hard targets.

\subsubsection{(d) Fine-tuning with hard targets:}
The student model is finally fine-tuned using the hard target dataset. This final step aims to refine the model's accuracy in relevant regions of the PES that were not adequately addressed during the soft targets learning phase. Since the hard targets are generated from a subset of the soft targets, the underlying atomic structures are essentially identical, differing only in their labels. This reduces the need for extensive retraining of the student model's descriptors, which have already been trained on the soft targets. To improve training efficiency, we fix the learned descriptors and optimize only the fitting network.

\section{Results}
\begin{figure}[t]
    \includegraphics[width=\textwidth]{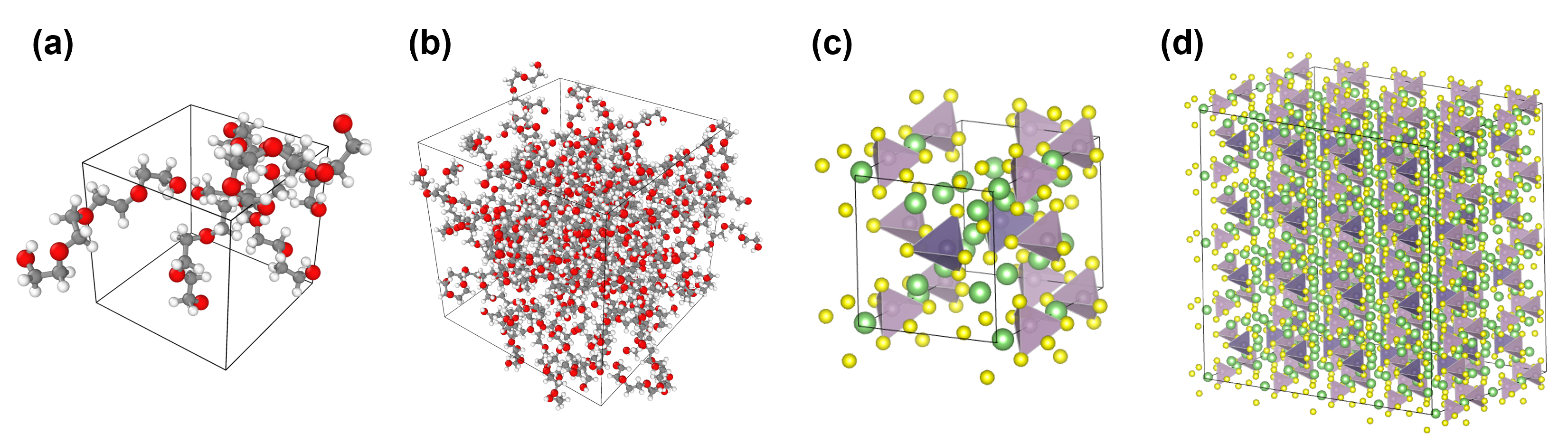}
    \caption{Atomic structures used in this work. (a) PEG system with five 4-mer molecules (155 atoms) for training data creation. (b) PEG system with one hundred 4-mer molecules (3100 atoms) for production MD simulations. (c) LGPS system containing two formula units (50 atoms) for training data creation. (d) LGPS system containing 64 formula units (1600 atoms) for production MD simulations.} 
    \label{fig:structures}
\end{figure}

To demonstrate the effectiveness of our KD framework, we apply it to polyethylene glycol (PEG), an organic solvent, and L$_{10}$GeP$_2$S$_{12}$ (LGPS), a lithium-ion conductor. We will present results detailing the accuracy and computational efficiency for these materials. The atomic structures used in this work, depicted in Figure~\ref{fig:structures}, were generated using the RadonPy~\cite{hayashi_radonpy_2022} for PEG and obtained from the Materials Project~\cite{jain_commentary_2013} (mp-696128) for LGPS.

\subsection{Accuracy verification for PEG}
To initiate our proposed framework, first, five initial structures were prepared by scaling the system depicted in Figure~\ref{fig:structures}(a) to densities of 1.120 g/cm$^3$ ± 5\% (ranging from 1.064 to 1.176 g/cm$^3$), based on the experimental value~\cite{hoffmann_densities_2021}.
MatterSim-v1.0.0-5M~\cite{yang_mattersim_2024} was employed as the teacher model to generate soft targets via NNP-MD simulations. These simulations were conducted for 150 ps with a timestep of 0.5 fs in the isothermal-isochoric ensemble at four different temperatures, ranging from 300 to 600 K in 100 K intervals. Atomic structures were extracted from the last 100 ps of the each trajectory every 50 fs to serve as soft targets. The resulting soft targets were split into training and validation datasets with an 8:2 ratio, yielding 16,000 training data points and 4,000 validation data points. This validation dataset was relabeled by DFT to assess the predictive accuracy of the NNP models.
DeepPot-SE (DP)~\cite{zhang_end--end_2018} implemented in the DeePMD-kit framework~\cite{wang_deepmd-kit_2018,zeng_deepmd-kit_2023} was used as the student model. The se\_e2\_a descriptor was employed with a cutoff radius of 6\AA. The descriptor and fitting networks were composed of (25, 50, 100) and (240, 240, 240) nodes with ResNet-like architectures~\cite{he_deep_2016}, respectively. These models were trained for 500,000 steps with a batch size of 4. The Adam optimizer~\cite{kingma_adam_2015} was used. The learning rate decayed from $1.0\times10^{-3}$ to $1.0\times10^{-8}$ with 5000 decay steps. In the fine-tuning step, the weights of the descriptor networks were fixed.
Dimensionality reduction for hard target selection was executed using the densMAP~\cite{narayan_assessing_2021}, and the number of hard targets was set to 1000. To validate the effectiveness of the structural feature-based screening method~\cite{matsumura_generator_2025}, we also generated hard targets using random sampling with different random seeds.
All DFT calculations were carried out using the Quantum ESPRESSO package~\cite{giannozzi_quantum_2009}. The Becke--Lee--Yang--Parr (BLYP) exchange-correlation functional~\cite{becke_density-functional_1988} was used. The Hartwigsen--Goedecker--Hutter (HGH) pseudopotential~\cite{hartwigsen_relativistic_1998} was employed with a cutoff of 100 Ry. To account for the dispersion, an empirical D3 dispersion correction~\cite{grimme_consistent_2010} was applied.

The mean absolute error (MAE) of forces against the DFT-labeled validation data for the student model trained solely on 16,000 soft targets was 0.204 eV/\AA, mirroring the teacher model's MAE of 0.202 eV/\AA. This close agreement confirms the successful knowledge transfer from the teacher to the student model. The MAE of the fine-tuned student model was 0.061 eV/\AA, indicating that the fine-tuning with hard targets further enhances the model's predictive accuracy.

\begin{table}[t]
    \caption{Comparison of PEG densities and self-diffusion coefficients at 298.15 K and 1 bar. Values in parentheses represent the percentage deviation from the experimental values. The teacher model is ``MatterSim-v1.0.0-5M''~\cite{yang_mattersim_2024}. ``This work'' refers to sutdent models generated using our KD framework. ``Random'' indicates hard target selection using random sampling, as opposed to the structural feature-based screening~\cite{matsumura_generator_2025}.}
    \label{tab:peg_results}
    \centering
    \begin{tabular}{l@{\hspace{0.3cm}}c@{\hspace{0.3cm}}c@{\hspace{0.3cm}}c}
        \toprule
        & \textbf{\# of} & \multirow{2}{*}{\textbf{Density}} & \textbf{Self-diffusion} \\
        \textbf{Method} & \textbf{hard} & \multirow{2}{*}{\textbf{(g/cm$^3$)}} & \textbf{coefficient} \\
        & \textbf{targets} &  & \textbf{(10$^{-6}$ cm$^{2}$/s)}\\
        \midrule
        Experiments~\cite{hoffmann_densities_2021}     & ---     & 1.120 & 0.297 \\
        \midrule
        OPLS4~\cite{lu_opls4_2021}                     & ---     & 1.113 (-1\%) & 0.022 (-93\%) \\
        QRNN~\cite{mohanty_development_2023}           & 344,654 & 1.128 (1\%) & 0.390 (31\%) \\
        GeNNIP4MD~\cite{matsumura_generator_2025}      & 9987    & 1.141 (2\%) & 0.235 (-21\%) \\
        MatterSim-v1.0.0-5M (zero-shot)~\cite{yang_mattersim_2024} & ---     & 0.743 (-34\%) & 9.676 (3158\%) \\
        \midrule
        This work (Soft targets only)                  & ---     & 0.994 (-11\%) & 0.775 (161\%) \\
        This work (Our proposal)                       & 1000    & 1.132 (1\%) & 0.334 (12\%) \\
        This work (Our proposal)                       & 500     & 1.031 (-8\%) & 0.741 (149\%) \\
        This work (Random, seed 0)                     & 1000    & 1.152 (3\%) & 0.170 (43\%) \\
        This work (Random, seed 1)                     & 1000    & 1.130 (1\%) & 0.288 (-3\%) \\
        \bottomrule
    \end{tabular}
\end{table}

Here, we examine the accuracy of the student model for reproducibility of density and self-diffusion coefficient by performing long-time NNP-MD simulations. The self-diffusion coefficient was estimated from the mean square displacement of all atoms. 
NNP-MD simulations were run for 21 ns with a timestep of 0.5 fs in the isothermal-isobaric ensemble at 298.15 K and 1 bar, and the properties were estimated from the last 20 ns of the trajectory.

Table~\ref{tab:peg_results} compares these values with experimental data~\cite{hoffmann_densities_2021} and MD-calculated values from a traditional empirical force field (OPLS4~\cite{lu_opls4_2021}) and NNP models (QRNN~\cite{mohanty_development_2023}, GeNNIP4MD~\cite{matsumura_generator_2025}, and MatterSim~\cite{yang_mattersim_2024}). 
Notably, the teacher model, MatterSim-v1.0.0-5M, significantly underestimates the density compared to experimental values. This discrepancy likely stems from the lack of dispersion corrections (D3 correction) in the training data, which compromises the accuracy of intermolecular interaction predictions. As a consequence, the self-diffusion coefficient is overestimated.
Meanwhile, the student model trained solely on soft targets demonstrates improved reproducibility of both density and self-diffusion coefficient compared to the teacher model. This improvement is likely due to the specialization of the learning process towards PEG, as focusing on a specific task has been shown to enhance learning efficiency in other studies~\cite{jiao_tinybert_2019,sanh_distilbert_2019}.
Furthermore, fine-tuning this student model with D3-corrected 1,000 hard targets generally leads to improved properties predictions. Notably, the fine-tuned student model on the hard targets selected by our proposed framework results in a highly accurate reproduction of the experimental density, within 1\%.
While the random sampling with seed 1 also shows great agreement with the experimental density, that with seed 0 significantly underestimates the self-diffusion coefficients, indicating a lack of robustness in the approach.
To assess the trade-off between the number of hard targets and accuracy, we evaluated performance using only 500 data points selected by our proposed framework. Although density reproducibility remained within 8\%, these results suggest that utilizing 1000 data points is required for achieving a desirable level of accuracy when modeling PEG.
Importantly, our framework achieves comparable accuracy to GeNNIP4MD, which trains its NNP from scratch. This demonstrates a significant gain in data efficiency, requiring 10x fewer hard targets.

Next, to evaluate the relative performance of our proposed framework, we compare it to existing KD-based NNP generation methods for MD simulations~\cite{zhang_dpa-2_2024,wang_pfd_2025}. 
Unlike our approach, these methods first fine-tune a teacher model on hard targets. Soft targets are then generated using the fine-tuned teacher model, and a student model is trained solely on these soft targets. To replicate this method, we conducted an additional test by the steps outlined below:

\begin{enumerate}
    \item Fine-tune the teacher model using 1000 hard targets, identical to the ``Random, seed 1'' dataset presented in Table~\ref{tab:peg_results}.
    \item Generate new soft targets by performing NNP-MD simulations with the fine-tuned teacher model. We denote these as ``Soft targets (Fine-tuned)'', while the original soft targets are referred to as ``Soft targets (Pre-trained)''.
    \item Train a student model on the Soft targets (Fine-tuned).
\end{enumerate}

\begin{figure}[t]
    \centering
    \includegraphics[width=0.8\textwidth]{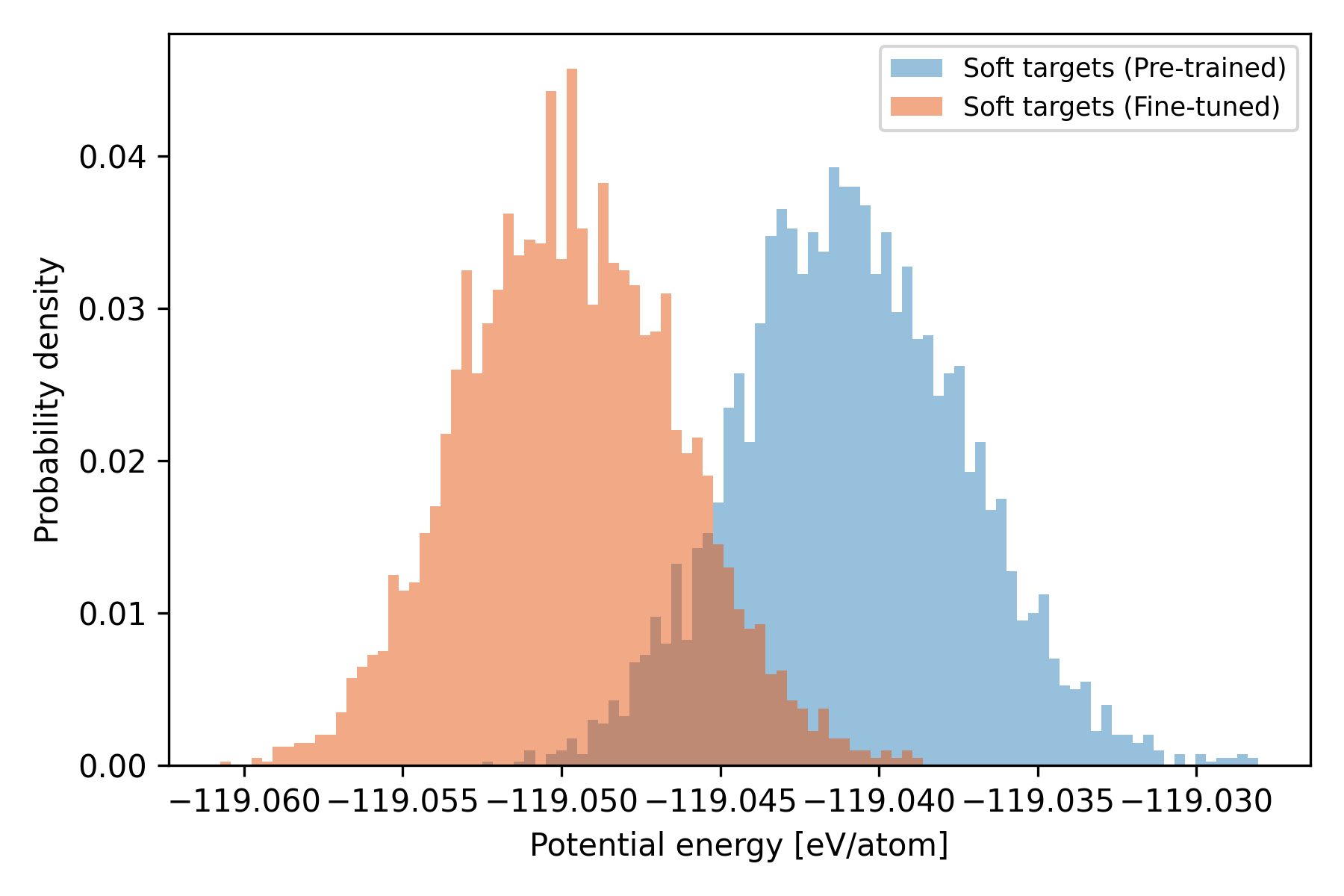}
    \caption{Energy histograms at 300 K for structures in the Soft targets (Pre-trained) (blue) and the Soft targets (Fine-tuned) (orange). These energies were relabeled by DFT as the ground truth. The distribution of the Soft targets (Fine-tuned) shifts towards lower values compared with that of the Soft targets (Pre-trained) due to the fine-tuning of the teacher model's PES.}
    \label{fig:histgram}
\end{figure}

As a result of the fine-tuning process, the fine-tuned teacher model achieved a force MAE of 0.030 eV/\AA~on the validation dataset. A subsequent production run using this teacher model yielded a density of 1.072 g/cm$^3$ and a self-diffusion coefficient of 3.89 $\times$ 10$^{-6}$ cm$^{2}$/s, confirming that the fine-tuning enhanced the non-fine-tuned teacher model.
A student model trained on the Soft targets (Fine-tuned) exhibited a force MAE of 0.063 eV/\AA, seemingly indicating satisfactory predictive accuracy. However, when this student model was used for MD simulations, the resulting density was 0.015 g/cm$^3$ and the self-diffusion coefficient was 6156.33 $\times$ 10$^{-6}$ cm$^{2}$/s, failing to reproduce experimental values.
To understand the underlying cause, we created energy histograms of the Soft targets (Pre-trained) and Soft targets (Fine-tuned), as shown in Figure~\ref{fig:histgram}. These histograms were relabeled by DFT as the ground truth.
From the figure, indeed, the Soft targets (Fine-tuned) exhibit an energy distribution shifted to the left compared to the Soft targets (Pre-trained), indicating insufficient sampling of data in the high-energy region. This demonstrates that fine-tuning of the teacher model increases energy barriers during MD simulations and hindering exploration in the high-energy region. This insufficient learning of structures in the high-energy region likely leads to unstable MD simulation, as reported in previous works~\cite{matsumura_generator_2025,yoshimoto_large-scale_2025}, and significant underestimation of density. Therefore, these results highlight the usefulness of our approach, which directly generates soft targets using a non-fine-tuned teacher model, enabling better sampling of high-energy configurations and contributing to improve the stability of MD simulations.

\subsection{Accuracy verification for LGPS}
We now present the results of applying our framework to LGPS. MatterSim-v1.0.0-5M~\cite{yang_mattersim_2024} was employed as a teacher model, and NNP-MD simulations were performed for 110 ps with a timestep of 0.5 fs in the isothermal-isochoric ensemble at five different temperatures, ranging from 300 to 1500 K in 300 K intervals. Structures were extracted from the last 100 ps of each trajectory every 50 fs as soft targets. The soft targets were then split into training and validation sets with an 8:2 ratio, resulting in 8000 training data points and 2000 validation data points. The projector augmented-wave (PAW) method~\cite{blochl_projector_1994} and the Perdew--Burke--Ernzerhof (PBE) exchange-correlation functional~\cite{perdew_generalized_1996} were employed for DFT calculation. All other settings were identical to those used for PEG.

\begin{figure}[t]
    \centering
    \includegraphics[width=0.75\textwidth]{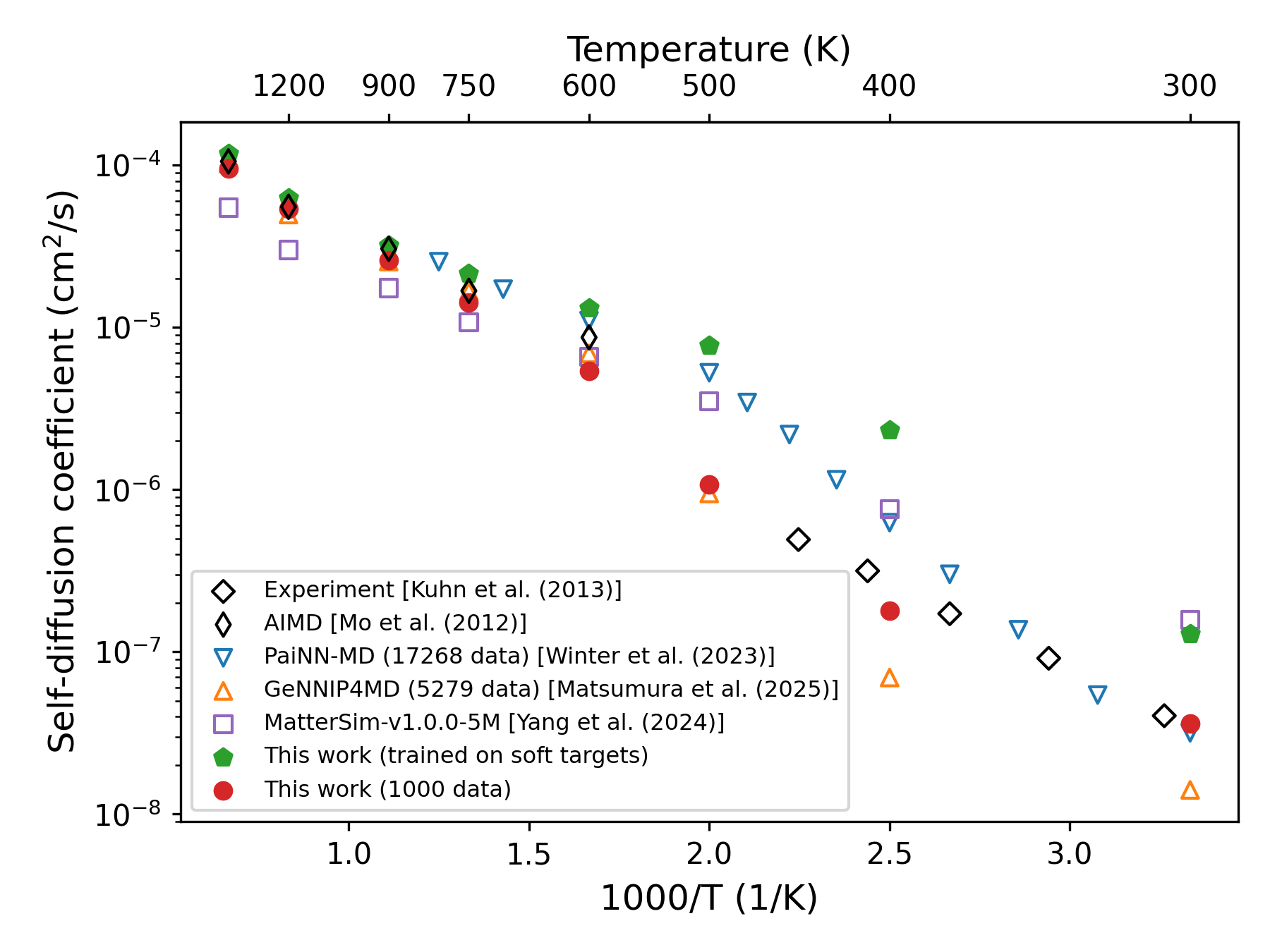}
    \caption{Arrhenius plot of lithium-ion self-diffusion coefficients for LGPS, obtained from experimental results~\cite{kuhn_tetragonal_2013}, AIMD simulations~\cite{mo_first_2012}, NNP-MD simulations using PaiNN model~\cite{winter_simulations_2023}, GeNNIP4MD~\cite{matsumura_generator_2025}, the teacher model (MatterSim-v1.0.0-5M)~\cite{yang_mattersim_2024} and our proposed DP model (trained on soft targets and fine-tuned on hard targets). Our DP model trained on solely soft targets overestimates them in the 300--500 K range. In contrast, the DP model fine-tuned with only 1000 hard targets demonstrates excellent agreement with AIMD and experiment values, highlighting the efficiency of our proposed framework.}
    \label{fig:difussion_coefficient_LGPS}
\end{figure}

We now present the results of our proposed framework for LGPS.
The student model, trained on soft targets, achieved a force MAE of 0.088 eV/\AA~on the DFT-labeled validation data, closely matching the teacher model's 0.085 eV/\AA~and demonstrating successful knowledge transfer. Fine-tuning with hard targets further improved the student model's accuracy, resulting in an MAE of 0.054 eV/\AA.
Figure~\ref{fig:difussion_coefficient_LGPS} presents the lithium-ion self-diffusion coefficients for LGPS. Winter et al.'s NNP-MD simulation\cite{winter_simulations_2023}, employing a PaiNN model~\cite{schutt_equivariant_2021} trained on 17,268 data points, accurately captures the non-linear temperature dependence of the diffusion coefficients, bridging the gap between the AIMD simulations~\cite{mo_first_2012} and experiments~\cite{kuhn_tetragonal_2013}. However, it overestimates the self-diffusion coefficient around 450 K. The DP-MD simulation of GeNNIP4MD, using a smaller dataset of 5279 data points and training DP from scratch, shows good agreement with AIMD values at high temperatures, similar to PaiNN-MD, but underestimates the self-diffusion coefficients at low temperatures. While our KD-based DP model trained solely on soft targets overestimated them at low temperatures, mirroring the behavior of the teacher model (MatterSim-v1.0.0-5M~\cite{yang_mattersim_2024}), fine-tuning with hard targets led to a significant improvement. This resulted in a much better agreement with experimental results, demonstrating the enhanced accuracy of our approach. Importantly, our framework achieved this level of accuracy using only 1/5 the data points required by GeNNIP4MD.

\subsection{Evaluation of computational performance}
This section presents a comparative analysis of the computational performance of NNP-MD simulations using the teacher model (MatterSim-v1.0.0-5M~\cite{yang_mattersim_2024}) and the student model (DP~\cite{zhang_end--end_2018}). We also compare the NNP generation time of our proposed framework against the AL-based NNP generation method~\cite{matsumura_generator_2025}.

\begin{figure}[t]
    \centering
    \includegraphics[width=0.8\textwidth]{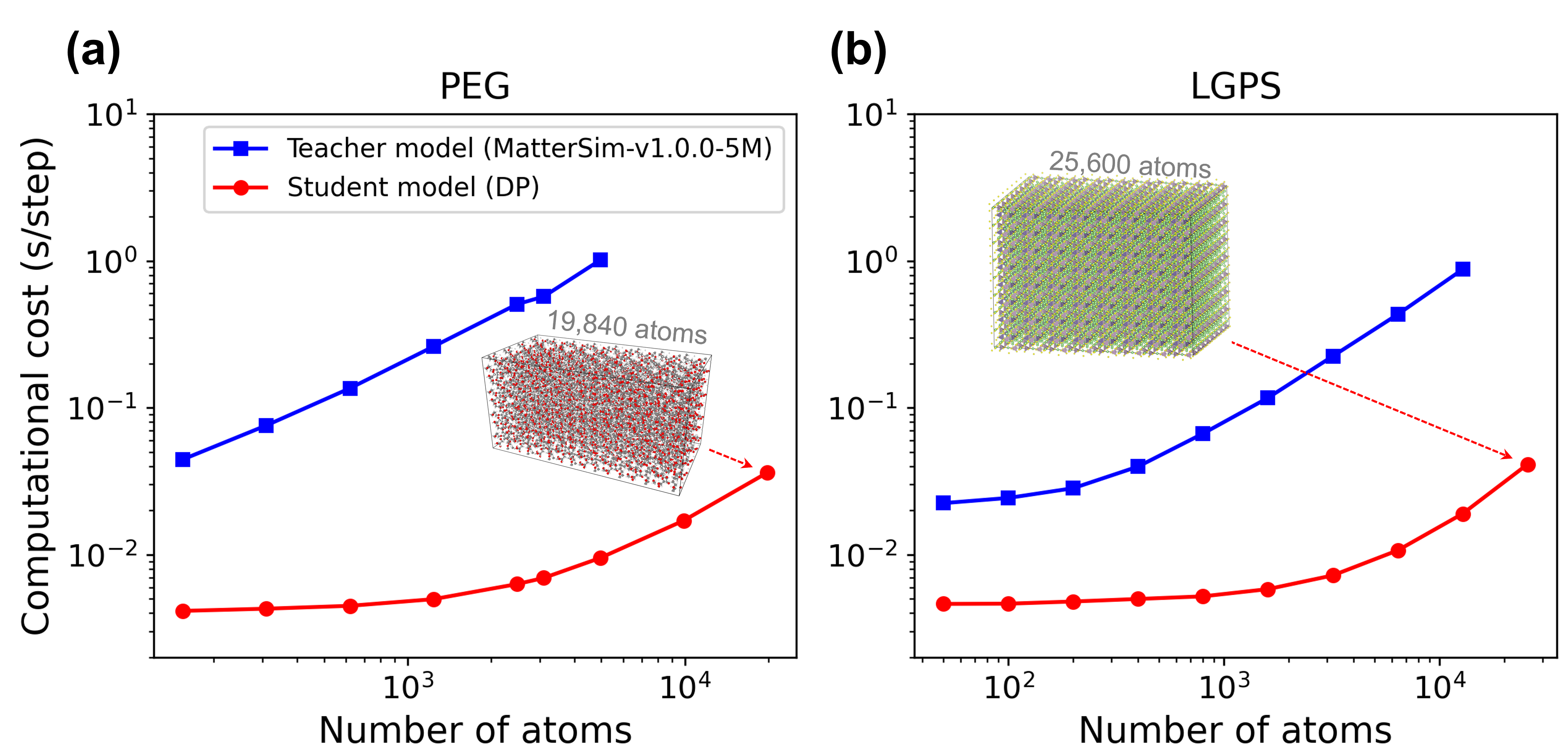}
    \caption{Computation time of teacher model (MatterSim-v1.0.0-5M~\cite{yang_mattersim_2024}) and student model (DP~\cite{zhang_end--end_2018}) for (a) PEG and (b) LGPS on an NVIDIA H100 80 GB GPU. For the PEG, the teacher model's high memory usage restricts simulations of PEG to a maximum of 4096 atoms. The student model enables simulations of much larger PEG systems (up to 19,840 atoms) and also achieves speedups of up to 107 times.}
    \label{fig:computation_time}
\end{figure}

To assess the computational time of NNP-MD simulation, we measured the average execution time of 1000-step NNP-MD simulations over five trials. These simulations were performed on an NVIDIA H100 Tensor Core 80 GB GPU and an Intel Xeon Gold 6430 CPU. As shown in Figure~\ref{fig:computation_time}, the student model achieves speedups of 10 to 106 times for PEG and 5 to 46 times for LGPS compared to the teacher model, highlighting the computational benefits of the student model. 
Specifically, for the 3100-atom (PEG) and 1600-atom (LGPS) systems used to calculate the physical properties in this work, we observed speedups of 82 times and 20 times, respectively.

Regarding NNP generation cost, all NNP generation processes were conducted on an NVIDIA Tesla V100 SXM2 16 GB GPU and an Intel Xeon E5-2698 v4 @ 2.20GHz CPU. Table~\ref{tab:nnp_generation_time} summarizes the NNP generation time, focusing on the elapsed time for structure sampling (soft target generation), NNP training, and DFT labeling (hard target generation), which collectively account for over 95\% of the total execution time. As detailed in the table, our KD-based framework enables NNP generation 1.9 times faster for PEG and 3.0 times faster for LGPS compared to the AL-based approach. This significant acceleration is primarily attributed to two factors: the elimination of iterative NNP retraining required in each loop of the AL process, and the reduction of computationally expensive hard-target generation.

\begin{table}[t]
    \caption{Comparison of NNP generation time for PEG and LGPS on an NVIDIA V100 GPU. ``GeNNIP4MD'' refers to the AL-based approach~\cite{matsumura_generator_2025}, while ``This work'' refers to our KD framework. Values in parentheses indicate speedup relative to ``GeNNIP4MD''.}
    \label{tab:nnp_generation_time}
    \centering
    \begin{tabular}{l@{\hspace{0.3cm}}l@{\hspace{0.3cm}}c@{\hspace{0.3cm}}c@{\hspace{0.3cm}}c@{\hspace{0.3cm}}c}
        \toprule
        & & \multicolumn{4}{c}{\textbf{Elapsed time (h)}} \\
        \cmidrule{3-6}
        \textbf{System} & \textbf{Method} & \textbf{Structure} & \textbf{DFT labeling} & \multirow{2}{*}{\textbf{Training}} & \multirow{2}{*}{\textbf{Total}}\\
         & & \textbf{sampling} & \textbf{(Hard target)} &  \\
        \midrule
        PEG & GeNNIP4MD~\cite{matsumura_generator_2025} & 38 & 101 & 121 & 260\\
        & This work & 123 & 8 & 7 & \textbf{138 (1.9x)} \\
        \midrule
        LGPS & GeNNIP4MD~\cite{matsumura_generator_2025} & 4 & 132 & 99 & 235 \\
        & This work & 45 & 25 & 8 & \textbf{78 (3.0x)} \\
        \bottomrule
    \end{tabular}
\end{table}

\section{Conclusion}
In this work, we have presented a novel KD framework for generating lightweight and high-accuracy NNPs tailored for MD simulations. Unlike existing KD methods that fine-tune a teacher model and subsequently sample low-energy configurations, our approach leverages a pre-trained universal NNP to explore high-energy regions, increasing the probability of including representative structures for improving the robustness of NNPs in the training dataset. The student model is initially trained on this dataset, and then refined through fine-tuning with high-accuracy DFT calculations on strategically selected data points. Results on PEG and LGPS demonstrate that our framework can generate NNPs that accurately reproduce physical properties while reducing the computational cost of NNP-MD simulations by up to two orders of magnitude compared to the teacher model. This framework offers a promising approach for accelerating materials simulations by enabling the efficient generation of accurate and computationally affordable NNPs. Future work will focus on extending this approach to more complex materials systems and establishing efficient strategies for soft target generation, thereby enabling even faster NNP generation.

\bibliographystyle{splncs04_modified}
\bibliography{library}

\end{document}